%% file: main.tex
\documentclass[review]{elsarticle}
\usepackage[top=1in,bottom=1in,left=1in,right=1in]{geometry}
\usepackage{latexsym, graphicx, epsfig, amsmath, amssymb, amsfonts}
\usepackage{amsmath}
\usepackage{amssymb}
\usepackage{graphicx}
\usepackage{subfig}
\usepackage{tabu}
\usepackage{algorithm}
\usepackage{algpseudocode}
\usepackage{adjustbox}
\usepackage{bbold}
\usepackage{bm}
\usepackage{textcomp}
\usepackage[thinc]{esdiff}
\usepackage[dvipsnames]{xcolor}
\usepackage{comment}
\usepackage{soul}
\usepackage{multirow}
\biboptions{sort&compress}

\newcolumntype{L}[1]{>{\raggedright\let\newline\\\arraybackslash\hspace{0pt}}m{#1}}
\newcolumntype{C}[1]{>{\centering\let\newline\\\arraybackslash\hspace{0pt}}m{#1}}
\newcolumntype{R}[1]{>{\raggedleft\let\newline\\\arraybackslash\hspace{0pt}}m{#1}}

\begin{document}
\begin{frontmatter}
\title{A Generative Modeling Framework for Inferring Families of Biomechanical Constitutive Laws in Data-Sparse Regimes}

\author{Minglang Yin\textsuperscript{ab\dag}}
\author{Zongren Zou\textsuperscript{c\dag}}
\author{Enrui Zhang\textsuperscript{c}}
\author{Cristina Cavinato\textsuperscript{d}}
\author{Jay D. Humphrey\textsuperscript{d}\corref{cor2}}
\cortext[cor2]{Corresponding author: jay.humphrey@yale.edu}
\author{George Em Karniadakis\textsuperscript{abc}\corref{cor1}}
\cortext[cor1]{Corresponding author: george\_karniadakis@brown.edu}
\cortext[]{\dag These authors contribute equally.}
\address[a]{Center for Biomedical Engineering, Brown University, Providence, RI}
\address[b]{School of Engineering, Brown University, Providence, RI}
\address[c]{Division of Applied Mathematics, Brown University, Providence, RI}
\address[d]{Department of Biomedical Engineering, Yale University, New Haven, CT}

\begin{abstract}

Quantifying biomechanical properties of the human vasculature could deepen our understanding of cardiovascular diseases. Standard nonlinear regression in constitutive modeling requires considerable high-quality data and an explicit form of the constitutive model as prior knowledge. By contrast, we propose a novel approach that combines generative deep learning with Bayesian inference to efficiently infer families of constitutive relationships in data-sparse regimes. Inspired by the concept of functional priors, we develop a generative adversarial network (GAN) that incorporates a neural operator as the generator and a fully-connected neural network as the discriminator. The generator takes a vector of noise conditioned on measurement data as input and yields the predicted constitutive relationship, which is scrutinized by the discriminator in the following step. We demonstrate that this framework can accurately estimate means and standard deviations of the constitutive relationships of the murine aorta using data collected either from model-generated synthetic data or ex vivo experiments for mice with genetic deficiencies. In addition, the framework learns priors of constitutive models without explicitly knowing their functional form, providing a new model-agnostic approach to learning hidden constitutive behaviors from data. 
\end{abstract}

\end{frontmatter}
\textbf{keywords:} Constitutive Relation, Uncertainty Quantification, Functional Prior, Generative Adversarial Network, Deep Operator Network, Bayesian Inference
\newpage

\section{Introduction}

Quantification of the mechanical properties of the aorta is central to understanding many aspects of cardiovascular health and disease. These properties stem from interactions and associations between many different structural proteins, glycoproteins, and other components of the aortic wall~\cite{karnik2003critical}. Malfunctions of these components may lead to structural changes at both micro- and macro-scales, which can be observed as changes in biomechanical properties as, for example, in conditions such as Marfan syndrome~\cite{humphrey2013possible}, Elhers-Danlos syndrome~\cite{byers2017diagnosis}, or many other diseases. Hence, deepening our understanding of vascular diseases necessitates the development of appropriate constitutive relations. 

Considerable progress has been made in modeling the mechanical behavior of the aorta~\cite{humphrey1995mechanics,fung2013biomechanics,simon1972reevaluation,vaishnav1972nonlinear,fung1967elasticity,fung1979pseudoelasticity,holzapfel2000new,baek2007theory}. Recently, various exponential models that account for fibers undulation and orientation have been proposed to describe constitutive behaviors of the healthy aorta, including two widely-adopted models, namely the Holzapfel-Gasser-Odgen (HGO or two-fiber family) model~\cite{holzapfel2000new} and the four-fiber family (4FF) model~\cite{baek2007theory}. These models contain an isotropic term (the neo-Hookean strain energy density) for the amorphous constituents (e.g., elastic laminae and aggregating glycosaminoglycans) and anisotropic terms (several exponential terms) for characterizing the fibrous extracellular matrix in different primary directions~\cite{ferruzzi2013biomechanical,masson2008characterization,jadidi2021constitutive}. Both two- and four-fiber family models describe experimental data well, the latter more so in some cases~\cite{jadidi2021constitutive}. 

Several challenges yet remain in estimating biomechanical properties of the aorta. First, data acquired from in vivo or ex vivo experiments are typically noisy and sparse, which may compromise the accuracy of parameter identification, often with convergence to a local rather than global minimum. To mitigate this, one could estimate better initial values in optimization and adopt more data and regularization terms. Recently, research using Bayesian inference provided an alternative solution~\cite{ninos2021uncertainty,eck2016guide,biehler2015towards,staber2018random,brewick2018uncertainty,flaschel2021unsupervised}. Particularly, Rego et al.~\cite{rego2021uncertainty} addressed this issue by developing a Bayesian framework for quantifying the local material properties of the ascending thoracic aorta. 
A more significant challenge relates to the need to specify a good function form for the constitutive models. Historically, proposing a new constitutive model is, in general, a particularly challenging task as it is time-consuming and requires deep physical insights. Related to this, most developed constitutive models are phenomenological, meaning that the models adopt terms to approximate the observed behavior and neglect or simplify microscopic details and dynamics; the HGO and 4FF models neglect fiber interactions across different directions. This challenge is increasingly pressing nowadays as 1) more meta- and bio-materials are created and discovered, and 2) they may be affected by many neglected factors, such as microstructure or genotype, leading to inaccurate predictions. Altogether, these two challenges collectively represent a hurdle to the development of accurate constitutive relations. 

Recently, more attention has shifted to data-driven constitutive modeling since only limited knowledge of functional forms is required while essential mathematical properties can be preserved. In this paper, we refer to such approaches as ``model-agnostic'' compared to knowing the explicit functional  forms (``model-specific''), noting that machine learning has greatly facilitated the development of this field~\cite{tac2022data,fuhg2022physics,thakolkaran2022nn,mozaffar2019deep,qu2021towards, huang2020learning,liu2020generic,liu2018estimation,liu2019estimation}. In particular, Masi et al~\cite{masi2021thermodynamics} proposed a thermodynamics-based neural network for learning material constitutive laws. The networks are informed with the two basic principles of thermodynamics in addition to stress-strain relationships to improve model generalization ability. Linka and colleagues~\cite{linka2021constitutive} proposed a neural network that takes the right Cauchy-Green tensor with a feature vector as input and predicts strain energy density. The network by $As'as$ et al.~\cite{as2022mechanics} preserves fundamental principles of mechanics and constraints, such as objectivity, consistency, or stability. Later, Linka et al.~\cite{linka2023new,linka2023automated} further developed a new class of neural networks for learning constitutive laws with enforced thermodynamic consistency, polyconvexity, and other physical constraints. 

Driven by the important roles of microstructure and genetics in dictating biomechanical properties, developing a data-driven constitutive model that incorporates these features has also received increasing attention~\cite{guo2021learning,linka2022predicting}.
Holzapfel and colleagues~\cite{holzapfel2021predictive} designed a neural network that predicts parameters of the HGO model based on histological analysis and images from second-harmonic generation. Later on, Linka et al.~\cite{linka2022predicting} extracted structural features in different vessel layers from two-photon imaging and fed these features into a physics-informed neural network. The network accurately predicts arterial mechanical properties based on structural features and opens up a way to quantitatively examine the relation between tissue microstructure and its macroscopic properties. However, most of the aforementioned works learn the constitutive relations based on destructive measurements or in a case-by-case approach whereas model retraining is needed if the material properties change. To address this limitation, Zhang et al.~\cite{zhang2022g2varphinet} proposed a generic framework to infer stress-strain relationships for murine aortas with diverse genotypes without assuming the function form \textit{a priori}. In particular, the network sampled in a latent space and picked the closest stress curves that match the measurement data collected from a new sample without retraining.

\begin{figure}
    \centering
	\includegraphics[width=0.7\textwidth]{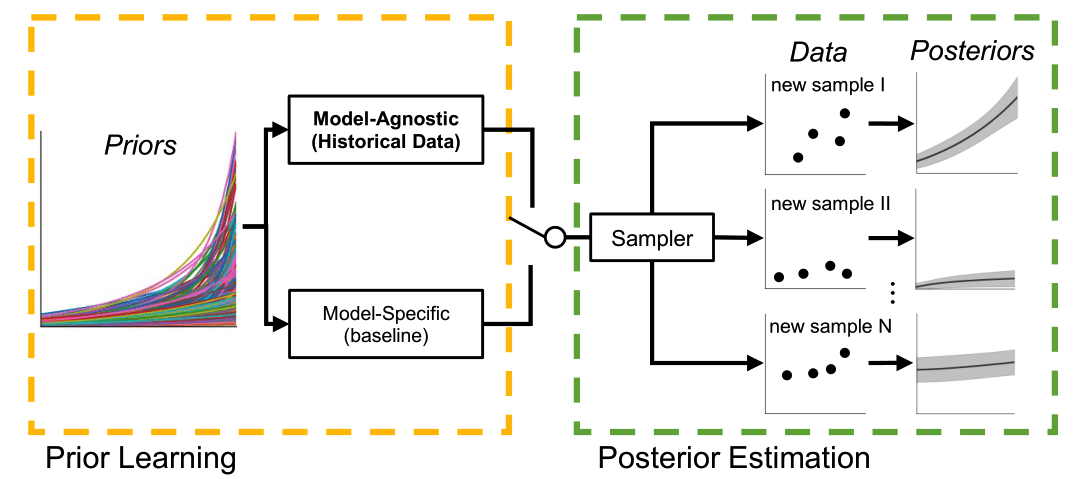} 
    \caption{\textbf{Schematic of inferring constitutive relationships with model-agnostic vs. model-specific approaches for families of materials.} Prior knowledge of a family of constitutive models can be expressed in a model-specific approach with its explicit function form or be informed in a model-agnostic deep learning model trained with historical data. Either of the models can be incorporated with a sampler to estimate the posterior for families of materials conditioned on new data collected from different samples. N samples can be collected from single or multiple genetic families. 
    }
    \label{fig:schematic}
\end{figure}


To address the aforementioned challenges, the present work develops a generative framework for inferring families of soft tissue constitutive relations in data-sparse regimes; an overview schematic is presented in Fig.~\ref{fig:schematic}. In contrast to previous model-specific approaches (with known function forms), our model-agnostic approach learns the constitutive relationships directly from historical data without knowing the functional forms. Inspired by the concept of a functional prior~\cite{meng2022learning}, the prior knowledge is implicitly captured by a generative adversarial network (GAN)~\cite{yang2020physics,yang2019adversarial,goodfellow2014generative}. After training, we combine the trained generator with Bayesian inference to efficiently estimate posteriors based on sparse measurements for multiple samples from single or multiple genetic families without model retraining (Fig.~\ref{fig:schematic}). Sampling methods, such as Markov Chain Monte Carlo, Hamiltonian Monte Carlo, or other related methods are adopted to estimate the posterior. As the output of Bayesian inference, a distribution of noise conditioned on new data is generated and transmitted to the GAN for estimating the means and standard deviations of stresses.

The paper structure is listed below: In Sec.~\ref{sec:method}, we briefly describe the 4FF model for modeling vascular biomechanics, the model-agnostic approach, as well as Bayesian inference framework for model-specific approaches. In Sec.~\ref{sec:result}, we report the inference accuracy for benchmarks in 1D and 2D, followed by performing posterior inference based on ex vivo experimental data. Finally, we conclude with a short discussion in Sec.~\ref{sec:discussion}.

\section{Methodology}
\label{sec:method}
In this section, we briefly describe the 4FF model in Sec.~\ref{sec:4ff}, GAN for learning priors in Sec.~\ref{sec:pi_gan}, Bayesian inference and methods for posterior estimation in Sec.~\ref{sec:hmc}, and the model-specific inference in Sec.~\ref{subsec:4ff_based}.

\subsection{Four-Fiber Model for Vascular Mechanics}
\label{sec:4ff}

We adopt the 4FF model as an illustrative phenomenological constitutive model for describing mechanics of the aorta and for generating training data~\cite{baek2007theory,jadidi2021constitutive,ferruzzi2013biomechanical}. As the functional form will not be an input to the generative network, adopting this constitutive model only serves as an example of the learning ability of our generative framework and demonstrates its potential for generalizing to other constitutive models. The 4FF model describes the aorta as hyperelastic, nonlinear, and anisotropic, the latter consistent with fiber-reinforcement in four diverse directions. The strain energy density is written as:
\begin{align}
    W & = \frac{\mu}{2}(I_{1} - 3) + \sum^{4}_{i=1}\frac{k^{i}_{1}}{4k^{i}_{2}}(\exp{(k^{i}_{2} (I_{4i} - 1)^{2})} - 1),
\end{align}
where $\mu$ is shear modulus and $k^{i}_{1, 2}, i=1,...,4$ are fiber modulus and exponential coefficients of the four fiber groups oriented along the axial ($i=1$), circumferential ($i=2$), and two symmetric diagonal directions ($i=3, 4$). The above parameters are usually calibrated using nonlinear regression methods based on experimental data~\cite{ferruzzi2013biomechanical,gavin2019levenberg}. $I_{1}$ is the first principal invariant and $I_{4i}$ denotes a pseudo principal invariant corresponding to the $i-$th fiber group:
\begin{align*}\label{eqn:fiberstrain}
    I_{1} &= tr\textbf{C}, \\
    I_{4i} &= \textbf{C}:\textbf{M}_{i} \otimes \textbf{M}_{i}, \quad i=1,..,4
\end{align*}
where \textbf{C} is the right Cauchy Green tensor. $\textbf{M}_{i}$ indicates the orientation of the $i$th locally parallel families of fibers ($i=1$, axial; $i=2$, circumferential, $i=3, 4$, symmetric diagonal at angle $\alpha$, where $\alpha$ is the angle of the diagonal families with respect to the axial direction.
Assume the material is incompressible ($J=1$) and no interaction between fibers, the Cauchy stresses (\bm{$\sigma$}) can be calculated by taking the partial derivatives of the energy density: 
\begin{equation}
\label{equ:stress}
        \bm{\sigma} = -p\bm{I} + 2\bm{F}\frac{\partial W}{\partial \bm{C}} \bm{F}^{T} = -p\bm{I} + \frac{\partial W}{\partial \bm{F}} \bm{F}^{T}
\end{equation}


\subsection{Model-Agnostic Approach for Learning Priors}
\label{sec:pi_gan}

\begin{figure}
    \centering
	\includegraphics[width=1.0\textwidth]{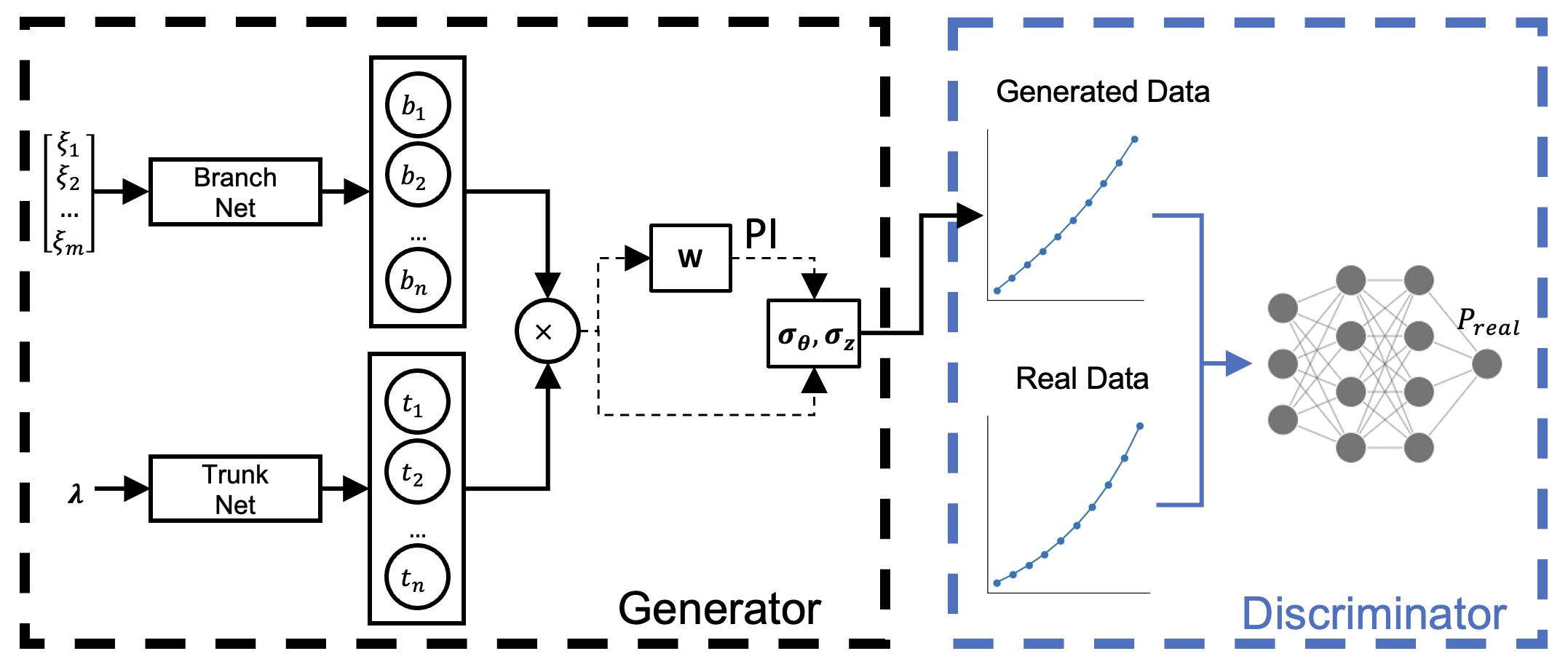} 
    \caption{\textbf{Generative adversarial networks (GANs) for learning prior knowledge of vascular constitutive laws.} The generator is a DeepONet with a vector of noise \pmb{$\xi$} and stretch \pmb{$\lambda$} as the input of the branch and trunk net. The generator predicts the corresponding stresses ($\sigma_{\theta}$ and $\sigma_{z}$) either directly or by taking the partial derivatives (physics-informed) of the intermediate output (strain energy density $W$). The generated stress field and real data will be transmitted to the discriminator for estimating their difference.}
    \label{fig:pi_gan}
\end{figure}

In this section, we briefly introduce how our model-agnostic approach is trained to learn constitutive relations as priors from historical data. GAN has shown promising performance in image generation due to its powerful ability for learning image priors. Herein, we adopt a GAN to learn the priors of constitutive relations. The model consists of a physics-informed deep operator network (PI-DeepONet) as the generator and a fully-connected neural network (FNN) as the discriminator. 

The architecture of GAN is presented in Fig.~\ref{fig:pi_gan}. The structure of the generator is inherited from DeepONet, which is a flexible and powerful neural network for learning mappings between functions~\cite{lu2021learning,yin2022interfacing,yin2022simulating,lin2021operator,zhang2022hybrid}. As shown in the left panel of Fig.~\ref{fig:pi_gan}, the network consists of a branch and a trunk net. The branch net takes a vector $\bm{\xi}$ of Gaussian random variables as input and yields a latent vector $\bm{b} = [b_{1}, b_{2}, ..., b_{n}]$, while the trunk net takes a vector of stretches \bm{$\lambda$} as the input and outputs a vector $\bm{t} = [t_{1}, t_{2}, ..., t_{n}]$ with the same dimensionality as \bm{$b$}. In the original framework of DeepONet, the network output takes an inner product of those two vectors, \bm{$b$} and \bm{$t$}, namely $\Sigma^{n}_{i=1}b_{i}(\bm{\xi})t_{i}(\bm{\lambda})$, to approximate the function of interest. 
In this work, $\bm{\xi}$ is the source of randomness for generative modeling and each instance of $\bm{\xi}$ corresponds to a specific function, which takes as input $\bm{\lambda}$ and outputs the quantity of interest. 

We note that the data for training the GAN in learning priors (shown in the left panel of Fig.~\ref{fig:schematic}), denoted as $\tilde{T}$, is different from the data used for downstream posterior estimation (shown in the right panel of Fig.~\ref{fig:schematic}), denoted as $D$. $\tilde{T}$ is a dataset containing the information from the prior of a family of models and each data/instance of $\tilde{T}$ describes a specific model, that is, a function that relates the stretch $\bm{\lambda}$ to the stress $\bm{\sigma}$. In contrast, $D$ is a dataset containing information from a specific model and each data/instance of $D$ describes a measurement, that is, a value of the stress $\bm{\sigma}$ measured at a specific stretch $\bm{\lambda}$. In other words, $\tilde{T}$ contains many models/functions while $D$ contains many measurements for one model/function. 

The generator outputs the strain energy of the arterial wall (Fig.~\ref{fig:pi_gan} left panel), such as the generators in Sec.~\ref{res:2d_syn}. Notice that the energy density has a tendency of exponential growth at large stretches, therefore, we adopt an exponential form for predicting $W$ as the model output after pointwise multiplication: $\mathcal{G}(\bm{\lambda}, \bm{\xi}) = \exp{(\Sigma^{n}_{i=1}b_{i}(\bm{\xi})t_{i}(\bm{\lambda}))} - 1$. Cauchy stress $\bm{\sigma}$ is computed according to Equ.~\ref{equ:stress}. Since we assume that the aorta is a thin wall by neglecting the radial stress we obtain:
\begin{equation}
    \sigma_{r}=-p+\frac{\partial W}{\partial \lambda_{r}} \lambda_{r} = 0,
\end{equation}
where $\lambda_{r} = 1/(\lambda_{\theta}\lambda_{z})$. $\frac{\partial W}{\partial \bm{\lambda}}$ is calculated using automatic differentiation in machine learning packages~\cite{abadi2016tensorflow}. The modification has two advantages: first, we empirically observe that such a modification increases the expressivity of networks for the exponential family; second, the results of linear combination $y$ are controlled within $O(10^{0})$. 

Such a technique is usually referred to as ``physics-informed'' learning, which is an emerging field where the network is trained and regularized using the governing equations or other prior physical knowledge~\cite{karniadakis2021physics,raissi2019physics,cai2022physics,zhang2022analyses,jin2021nsfnets,reyes2021learning,zhang2020physics,yin2021non,liu2020generic,zou2023hydra, chen2023leveraging, goswami2022physics,wang2021learning}. As shown in Fig.~\ref{fig:pi_gan}, the generator can also directly predict stresses, as demonstrated in Sec.~\ref{res:1d} and Sec.~\ref{res:ex-vivo}. 
Directly predicting stresses facilitates such operation as stresses in various directions grow at different rates. 
As for the discriminator $\mathcal{D}$ (Fig.~\ref{fig:pi_gan} right panel), we choose FNN as the network structure whose size will be introduced in Sec.~\ref{sec:result}.
The generator $\mathcal{G}$ aims at approximating the data distribution, denoted as $P_{T}$, as accurately as possible whereas the discriminator $\mathcal{D}$ is trained to distinguish the distance between real data and data generated from $\mathcal{G}$. The two parts are trained adversarially until the networks reach equilibrium~\cite{goodfellow2014generative}. Quantitatively, the generator aims at minimizing:
\begin{equation}
    L_{\mathcal{G}} = -\mathbb{E}_{ \bm{\xi} \sim P_{\bm{\xi}}}[\mathcal{D}(\mathcal{G}( \bm{\lambda}, \bm{\xi}))],
\end{equation}
where $P_\xi$ denotes the density function of $\xi$, and the discriminator is trained to minimize:
\begin{equation}\label{eq:loss_discriminator}
    L_{\mathcal{D}} = \mathbb{E}_{\bm{\xi} \sim P_{\bm{\xi}}}[\mathcal{D}(\mathcal{G}(\bm{\lambda}, \bm{\xi}))] - \mathbb{E}_{T \sim P_{T}}[\mathcal{D}(T)] + \beta_{reg}\mathbb{E}_{\hat{T} \sim P_{\hat{T}}}(||\nabla_{\hat{T}} \mathcal{D} (\hat{T})||_{2}-1)^2,
\end{equation}
where $P_{\hat{T}}$ denotes the density function induced by sampling uniformly on the interpolation lines between $T$ and $\mathcal{G}(\bm{\lambda}, \bm{\xi})$. $T$ denotes a finite representation of a model from the target family. We note that $\mathbb{E}_{T \sim P_{T}}[\mathcal{D}(T)]$ in practice is approximated by $\frac{1}{N}\sum_{i=1}^N \mathcal{D}(T_i)$, where $T_i, i=1,...,N$ are the data of $\tilde{T}$ and $N$ is the number of training data for the GAN.
Note, too, that the first two terms in Eq.~\eqref{eq:loss_discriminator}, $\mathbb{E}_{\bm{\xi} \sim P_{\bm{\xi}}}[\mathcal{D}(\mathcal{G}(\bm{\lambda}, \bm{\xi}))] - \mathbb{E}_{T \sim P_{T}}[\mathcal{D}(T)]$, can be interpreted as the negative of the Wasserstein-1 distance between the learned distribution and the real one \cite{yang2020physics} that enhances the ability of $\mathcal{D}$ to discriminate generated data $\mathcal{G}(\bm{\lambda}, \bm{\xi})$ with real data $D$, while the last term serves as a gradient penalty term to regularize the training dynamics. In this work, the gradient penalty coefficient $\beta_{reg}$ is chosen as 0.1. More details can be found in~\cite{meng2022learning}. 

\subsection{Bayesian Inference}
\label{sec:hmc}

As mentioned in the previous section, training of a GAN yields a generator that maps the latent multivariate random variable with tractable distribution, $\bm{\xi}$, to $\bm{\sigma}(\bm{\lambda};\bm{\xi})$. 
After the training of the GAN, the goal is to estimate the posterior distribution of $\bm{\xi}$ conditioned on data $D:=\{(\bm{\lambda}_i, \bm{\sigma}_i)\}_{i=1}^N$, i.e. $P(\bm{\xi}|D)$, where $N$ denotes the number of measurements. Bayes's theorem states that
\begin{equation}\label{eq:posterior}
    P(\bm{\xi}|\textit{D}) = \frac{P(\textit{D}|\bm{\xi})P(\bm{\xi})}{P(\textit{D})} \simeq P(\textit{D}|\bm{\xi})P(\bm{\xi})
\end{equation}
where $\simeq$ denotes equality up to a constant since $P(D)$ is non-tractable. $P(\textit{D}|\bm{\xi})$ is the likelihood (probability of $D$ conditioned on $\bm{\xi}$) defined as:
\begin{equation}\label{eq:likelihood}
    P(\textit{D}|\bm{\xi}) = \prod_{i=1}^{N}\frac{1}{\sqrt{2\pi \epsilon^2}}\exp(-\frac{(\bm{\sigma}_{\mathcal{G}}(\bm{\lambda}_i;\bm{\xi}) - \bm{\sigma}_i)^2}{2\epsilon^2}),
\end{equation}
with $\bm{\sigma}_{\mathcal{G}}(\bm{\lambda};\bm{\xi})$ is the generator prediction at $\bm{\lambda}$ given $\bm{\xi}$, $\epsilon$ the standard deviation of the data, and $\bm{\sigma}_i, i=1,...,N$ are the measurements. If stresses are available in multiple directions are available (e.g. circumferential and axial stresses), the new likelihood is simply the joint distribution of those variables assuming the data are independent. The prior distribution of $\bm{\xi}$, $P(\bm{\xi})$, is chosen to be a multivariate Gaussian in both the training of the GAN and the downstream inference:
\begin{equation}
    P(\bm{\xi}) = (\frac{1}{\sqrt{2\pi}})^{d_{\bm{\xi}}}\exp{-\frac{||\bm{\xi}||^{2}}{2}},
\end{equation}
where $d_{\bm{\xi}}$ denotes the dimensionality of $\bm{\xi}$.

Posterior estimation could be performed in various ways~\cite{psaros2023uncertainty, zou2022neuraluq}, including variational inference, Laplace approximation, or others. In this work, we choose Markov Chain Monte Carlo (MCMC) methods to sample from the un-normalized distribution in Eq.~\eqref{eq:posterior}. Specifically, Hamiltonian Monte Carlo (HMC) \cite{neal2011mcmc} and its variants \cite{hoffman2014no} are employed as the sampling methods to obtain samples of $\bm{\xi}$ subject to $P(\bm{\xi}|D)$, which are then fed to the generator to estimate $\bm{\sigma}$, namely $\bm{\sigma}_{\mathcal{G}}(\bm{\lambda};\bm{\xi})$. We refer to the method described in this section as \textit{GAN-based} since the stochastic model for Bayesian inference is a GAN.

\subsection{Model-Specific Bayesian Inference}\label{subsec:4ff_based}

We can perform Bayesian inference to the 4FF model by treating the model parameters as random variables and establishing the same likelihood distribution as in Eq.~\eqref{eq:likelihood}. We denote the model parameters by $\Theta = \{\mu, k_{1,2}^i, i=1,...,4\}$ and the model $\bm{\sigma} = \bm{\sigma}(\bm{\lambda};\Theta)$. Similarly, given data of $\bm{\sigma}$ and $\bm{\lambda}$, $D$, the problem now is to estimate the posterior distribution $P(\Theta|D)\simeq P(D|\Theta)P(\Theta)$. We choose the same sampling methods, HMC and its variants, for posterior estimation. We refer to this approach as \textit{4FF-based} method. We note that compared to standard nonlinear regression methods based on the 4FF model, which often provide deterministic inferences of the model parameters, our approach treats the model parameters as random variables and therefore focuses on their predictive distributions.

We further remark that unlike the GAN-based method, the 4FF-based method requires selection of a prior distribution of the model parameters $\Theta$, which is crucial in Bayesian inference but remains a challenge in many problems arising in scientific machine learning~\cite{psaros2023uncertainty}. In contrast, the GAN-based method treats the prior distribution of $\bm{\xi}$, $P(\bm{\xi})$, as a hyperparameter in the training of GANs, hence $P(\bm{\xi})$ is fixed, which simplifies the following Bayesian inference. 

Prior distributions of model parameters in the 4FF-based method are shown in Table~\ref{table:case1_param} and will be discussed in Sec.~\ref{sec:result}; they are independent uniform distributions, which prevents direct deployment of HMC and its variants as a uniform distribution has discontinuous density function. This issue can be easily resolved by changing of variables and letting the transformed random variables have smooth density functions. Interested readers are directed to~\ref{app:A} for more details.

\section{Results}
\label{sec:result}

In this section, we illustrate the performance of our framework with several examples where stress measurements are sampled from synthetic and ex vivo experiments. We set the training epochs of GAN as 100,000 and use ADAM~\cite{kingma2014adam} as the optimizer. All of the networks are trained on a NVIDIA GeForce RTX 2060 GPU.

\subsection{Case I: 1D model, a pedagogical example}
\label{res:1d}

\begin{table}
\begin{center}
\begin{tabular}{|c c c c|} 
\hline
Parameters & Unit & Base Values & Variations\\ 
\hline
$\mu$ &  $kPa$  & 8.633 & [0.1, 5.0] \\ 
\hline
$k_{1}^{1}$ &  $kPa$   & 9.34 & [0.1, 5.0] \\
\hline
$k_{2}^{1}$ &  -  & 0.137 & [0.01, 2.0] \\
\hline
$k_{1}^{2}$ &  $kPa$    & 4.447 & [0.1, 5.0] \\ 
\hline
$k_{2}^{2}$ &  -  & 0.046 & [0.01, 2.0]\\
\hline
$k_{1}^{3, 4}$ &  $kPa$   & 0.015 & [0.1, 5.0]\\
\hline
$k_{2}^{3, 4}$ & -  & 1.193 & [0.01, 2.0]\\ 
\hline
$\alpha$ &  $^{\circ}$  & 28 & [0.1, 2.0]\\
\hline
\end{tabular}
\end{center}
\caption{ \textbf{4FF parameters and their variation range.} All of the training data are generated with parameters independently and uniformly sampled in the ranges whose upper/lower bounds are computed as the ``Base Values'' multiplying the upper/lower bounds of the ``Variations''. The ``Base Values'' are adopted from reference~\cite{ferruzzi2013biomechanical}. Parameter unit is presented in the second column.}
\label{table:case1_param}
\end{table}

We first study the feasibility of applying our GAN-based method for learning the 4FF model in a pedagogical 1D example. To generate training data for the GAN, we first fix the axial stretch $\lambda_{z}=1.44$ and sample the parameters of the 4FF model from independent uniform distributions, whose ranges are computed using  Table~\ref{table:case1_param}. The upper/lower bounds are computed as the ``Base Values'' multiplying the upper/lower bounds of the ``Variations''. The ``Base Values'' are adopted from~\cite{ferruzzi2013biomechanical} with a relatively wide variation. The problem is to mimic an empirical observation~\cite{ferruzzi2018combining,masson2008characterization}, where the abdominal aorta in a mouse preserves its length over a cardiac cycle, meaning that the in vivo axial stretch is nearly a constant.

\begin{table}
\begin{center}
\begin{tabular}{|c c c c c|} 
\hline
Name & Type & Structure & Input & Output \\ 
\hline
Branch Net (Generator) & FNN & [50, 64, 64, 64, 50] & $\bm{\xi}$ & t \\ 
\hline
Trunk Net (Generator) & FNN & [1, 64, 64, 64, 50] & $\lambda_{\theta}$ & b\\
\hline
Discriminator & FNN & [15, 64, 64, 64, 1] & $\sigma_{\theta}$ & $y_{dis}$ \\
\hline
\end{tabular}
\end{center}
\caption{ \textbf{Structure and size of the GAN in Case I.} For example, [50, 64, 64, 64, 50] represents a three-layer network that has 50-dimension input and 50-dimension output. Each hidden layer has a width of 64 neurons. FNN: fully-connected neural networks. }
\label{table:case1_netstruct}
\end{table}

In Table~\ref{table:case1_netstruct}, we tabulate more details of our GAN. All of the networks in case I are chosen as  fully-connected neural networks (FNNs) with three hidden layers. Each hidden layer contains 64 neurons. We adopt DeepONet~\cite{lu2021learning} as the generator whose branch net takes a vector of random variable $\bm{\xi}\in \mathbb{R}^{50}$. The trunk net takes the circumferential stretch ($\lambda_{\theta}$) as input. The generator predicts the circumferential stress ($\sigma_\theta$) formulated as $u_{\mathcal{G}} = \exp{(\textbf{bt})} - 1$, where \textbf{b} and \textbf{t} denotes the output of the branch and trunk net.

\begin{figure}
    \centering
	\includegraphics[width=1.0\textwidth]{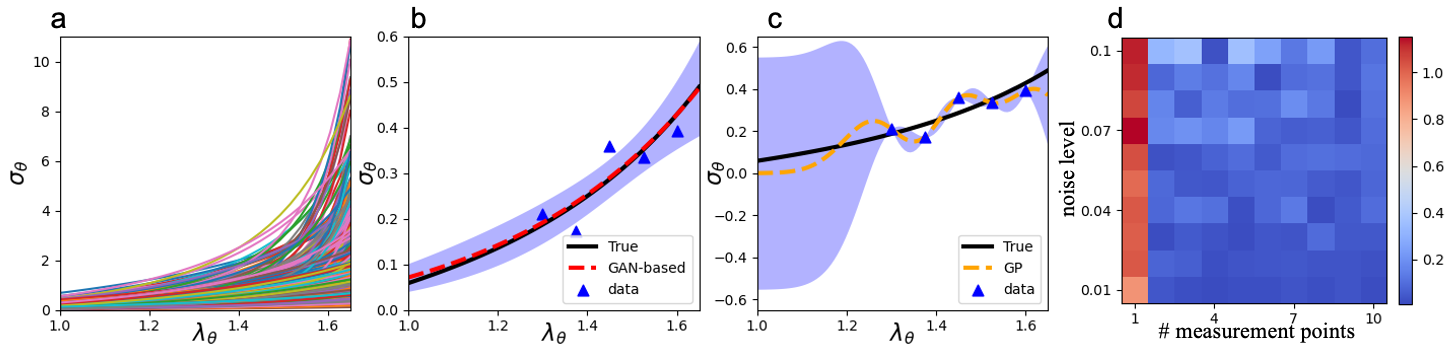} 
    \caption{\textbf{Comparison of FP (prior-informed model) and GP (Gaussian process, prior-uninformed model) for the constitutive model in 1D.} a): random samples of $\sigma_{\theta}$. Mean and standard deviations are computed from b) functional priors and c) Gaussian processes. d) 1D Prediction errors at different noise scales and the number of points. The color represents the magnitude of error defined in~\eqref{equ:err}. Stress unit: 0.1 MPa.}
    \label{fig:1D_comp}
\end{figure}

Fig.~\ref{fig:1D_comp}(a) presents samples that contain prior knowledge of the constitutive relationship in a training batch with $\sigma_{\theta}$ ranges from $O(10^{-1})$ to $O(10^{1})$. The training epochs are set as 100,000 with the ADAM optimizer. Upon completion of training, the objective is to infer the posteriors of circumferential stresses based on sparse and noisy measurements. We first draw five measurements of $\sigma_{\theta, 4FF}$ from $\lambda_\theta \in [1.3, 1.6]$ as measurement data, which are calculated from the 4FF model with parameter values set as the mean in Table~\ref{table:case1_param}. To test the model performance conditioned on noisy measurements, we contaminate $\sigma_{\theta, 4FF}$ with noise $e$ that satisfies a Gaussian distribution with its mean and variance at 0 and 1; the noise scale $\epsilon$ is set as 0.1. Hence, the measurements $\sigma_{\theta}$ are computed as:
\begin{equation}
    \sigma_{\theta} := \sigma_{\theta, 4FF} + \epsilon e, e \sim \mathcal{N}(0, 1),
\end{equation}
Fig.~\ref{fig:1D_comp}(b) presents the posterior inference from our framework (red dashed line), which matches well with the true data (black line). The relative error $\epsilon_{fp}$ is approximately at 0.02 when defined as follows:
\begin{equation}
\label{equ:err}
    \epsilon = \frac{|| \bar{\sigma}_{pred} - \bar{\sigma}_{true} ||}{|| \bar{\sigma}_{true} ||}.
\end{equation}
The means and standard deviations are averaged over 1,000 predictions from the generator based on noise vectors sampled from No-U-Turn. The shaded area indicates two standard deviations. As a comparison, Fig.~\ref{fig:1D_comp}(c) presents the means and standard deviations from Gaussian process (GP) regression. GP does not incorporate any prior knowledge of the 4FF model, leading to predictions with a higher error ($err = 0.22$). Notice that, given only five measurement points, the standard nonlinear regression fails since the system is under-determined: the number of variables is larger than the number of available data. Next, we investigate the impact of noise and the number of measurements on the model performance. It is clear from Fig.~\ref{fig:1D_comp}(d) that more measurement points with a lower noise level ($\epsilon$) could lead to more accurate predictions whereas fewer points with larger noise deteriorate the accuracy. 

\subsection{Case II: 2D Synthetic Data}
\label{res:2d_syn}

\begin{table}
\begin{center}
\begin{tabular}{|c c c c c|} 
\hline
Name & Type & Structure & Input & Output \\ 
\hline
Branch Net (Generator) & FNN & [100, 64, 64, 64, 50] & $\bm{\xi}$ & t \\ 
\hline
Trunk Net (Generator) & FNN & [2, 64, 64, 64, 50] & [$\lambda_{\theta}$, $\lambda_{z}$] & b\\
\hline
Discriminator ($\sigma_{\theta}$ and $\sigma_{z}$) & FNN & [1250, 250, 250, 250, 1] & [$\sigma_{\theta}$, $\sigma_{z}$] & $y_{dis}$ \\
\hline
\end{tabular}
\end{center}
\caption{ \textbf{Structure and size of the GAN in Case II where $\bm{\lambda} = \lambda_{\theta}, \lambda_{z}$ for the 2D problem.} FNN: fully-connected neural networks. }
\label{table:case2_netstruct}
\end{table}

In this section, we examine the performance of our GAN-based method in inferring arterial constitutive relationships using 2D synthetic data. All of the training and testing data are sampled and generated from the parameter setting in Table~\ref{table:case1_param}. We adopt a larger network structure shown in Table~\ref{table:case2_netstruct}. In the training stage, we inform the GAN with $\sigma_{\theta}$ and $\sigma_{z}$ as training data that are sampled at stretches ranging on a 2D grid $(\lambda_{\theta}, \lambda_{z}) \in [1.0, 1.65]\times [1.0, 1.65]$. 
Branch net input $\xi$ is a vector of 50 random variables sampled from multivariate Gaussian distribution in the training stage. The generator predicts the energy $W$ at a given stretch state and calculates the biaxial stresses $\sigma_{\theta,z}$ according to Eq.~\eqref{equ:stress} in the following step. The stresses are estimated on a grid of 25 by 25. Then, we concatenate the predicted $\sigma_{\theta}$ and $\sigma_{z}$ as the input of the discriminator for calculating the Wasserstein distance between the learned distribution and the real one. We systematically examine the same trained model on inferring 2D stress surfaces on random sampled points (Sec.~\ref{res_2:2d}), equi-stretch points (Sec.~\ref{res_2:equi}), partial data (Sec.~\ref{res_2:partial}), and out-of-distribution data (Sec.~\ref{res_2:ood}). We also study the effect of the number of training data for the GAN and the effect of locations of the measurements in posterior inference; see~\ref{app:B} and \ref{app:C}, respectively.

\subsubsection{Inference with Randomly Sampled Points}
\label{res_2:2d}

\begin{figure}
    \centering
	\includegraphics[width=1.0\textwidth]{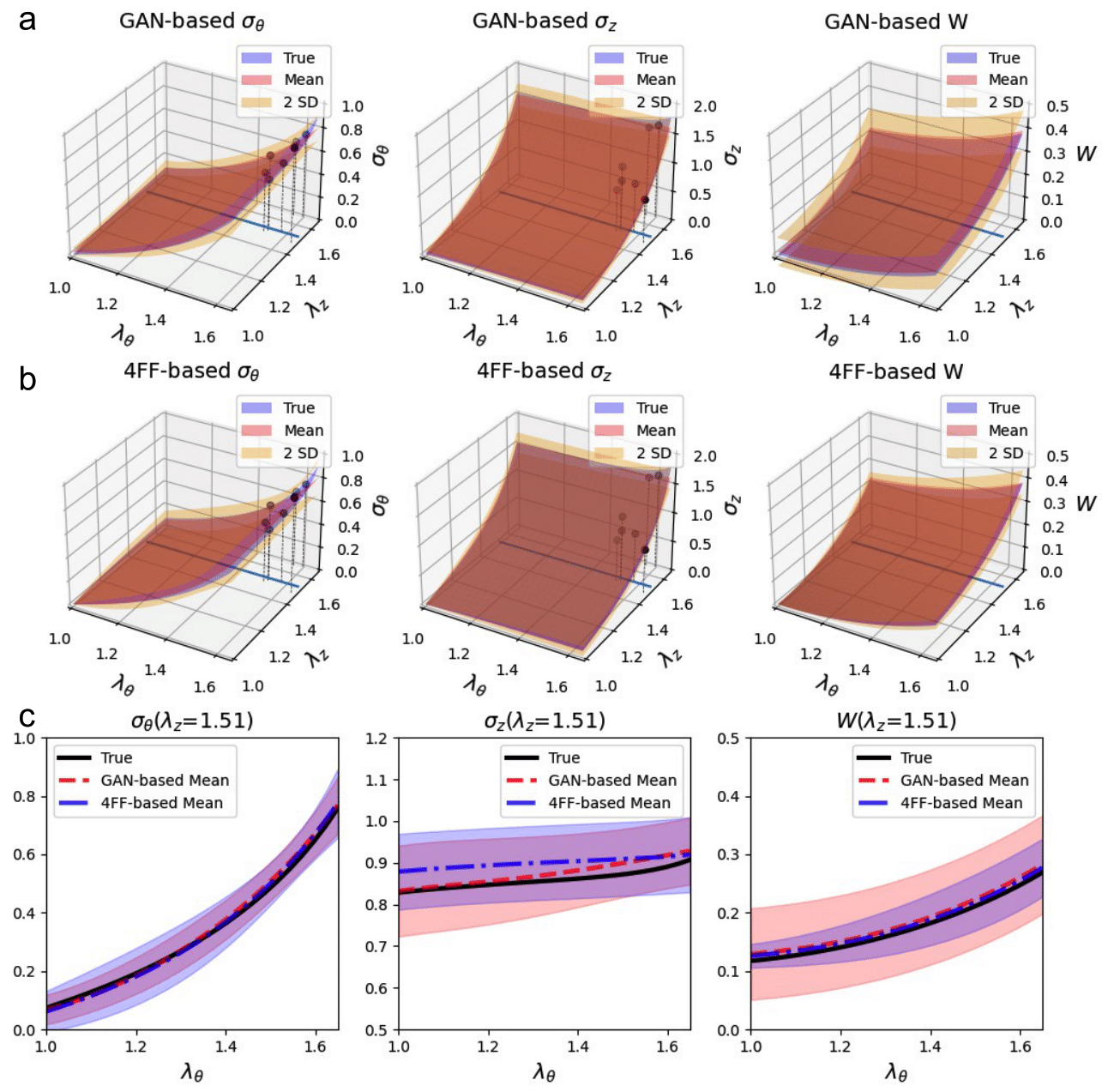} 
    \caption{\textbf{Posterior estimation of constitutive relationships from (a) FP and (b) BI based on 7 measurement points}. The mean (red) and 2 SD (orange) are plotted with the true data (blue) for $\sigma_{\theta}$, $\sigma_{z}$, and W from columns 1-3. Inference results when $\lambda_{z}=1.51$ are plotted against true results with uncertainty bounds. The measurement points containing $\sigma_{\theta}$ and $\sigma_{z}$ are generated from 4FF model. Stress unit: 0.1 MPa.}
    \label{fig:2d_surf_sc_sa}
\end{figure}

We examine the model performance given randomly sampled points that contain full stress information ($\sigma_{\theta}$ and $\sigma_{z}$). We first draw seven noisy measurements of $\sigma_{\theta}$ and $\sigma_{z}$ at high stretches $(\lambda_{\theta}, \lambda_{z}) \in [1.4, 1.6]\times [1.4, 1.6]$, where the noisy measurements are generated by the 4FF model with parameters at the base values in Table~\ref{table:case1_param}. Next, we adopt the No-U-Turn algorithm to sample 1,000 instances of $\bm{\xi}$ conditioned on the stress measurements. 
We then feed the sampled $\bm{\xi}$ into the generator to generate 1,000 different $\sigma_{\theta,z}$ for estimating the mean and standard deviations.

The top row in Fig.~\ref{fig:2d_surf_sc_sa} shows the inference results from the GAN-based method versus the true data (blue surface). The red surfaces, representing the mean estimation of $\sigma_{\theta, z}$ and $W$, match well with the true data bounded by standard deviations. The black dots in the first and second columns in (a) and (b) are the measured data. We set the noise level $\epsilon$ as 0.1. The results from 4FF-based inference (the second row) are the baseline model as it represents the best possible results based on the same measurements given the functional form. We observe a quantitative consistency in performance between the GAN-based and 4FF-based methods. To further compare these methods, we plot the inference results versus the true solution at $\lambda_{z}=1.51$ in the bottom row: the estimated means from the GAN-based and 4FF-based methods agree well with similar uncertainty regions at different stretches. However, the uncertainty in estimated energy ($W$) from the GAN-based is larger than that from the 4FF-based, possibly because the known functional form helps to regularize the network, yielding a reduced uncertainty as indicated by the blue-shaded region. Also, some estimations for $\sigma_{\theta}$ from the GAN-based method are non-physical at the region close to $\lambda_{\theta, z}=1$. 
Notice that the posterior inference is based on only seven noisy measurements whereas the standard nonlinear regression is under-determined in this scenario as there exist eight unknown parameters. Overall, results in Fig.~\ref{fig:2d_surf_sc_sa} show that the GAN-based method has a comparable performance with a model-specific approach (4FF-based) in inferring 2D constitutive surfaces. Moreover, the proposed framework is able to predict the energy density given that no prior knowledge of the energy profile is informed to the network in training.

\subsubsection{Inference with equi-stretch points}
\label{res_2:equi}

\begin{figure}
    \centering
	\includegraphics[width=1.0\textwidth]{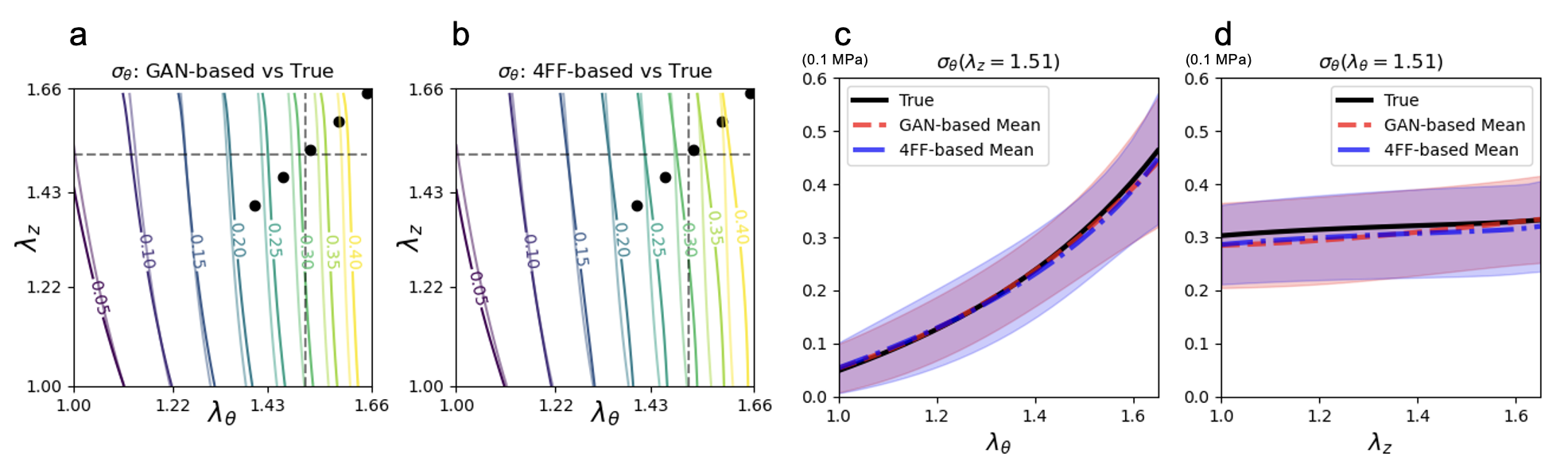} 
    \caption{\textbf{Model inference based on equi-stretch points.} True $\sigma_\theta$ is plotted against inference results from GAN-based and 4FF-based methods. Standard deviation is indicated by the shaded areas. Stress unit: 0.1 MPa.}
    \label{fig:2d_equi_stretch}
\end{figure}

Next, we examine the model performance based on equi-stretch measurements ($\lambda_{z}=\lambda_{\theta}$). In a standard nonlinear parameter fitting, these points are sometimes excluded given certain specific functional forms of the strain energy W. We summarize the stress inference using the GAN-based method with equi-stretch points in Fig.~\ref{fig:2d_equi_stretch}. We use the same GAN model in Sec.~\ref{res:2d_syn}. The left two panels (a and b) show contour lines of $\sigma_{\theta}$ inferred from the two inference models separately, where the level sets of true data (dashed line) are presented in a lighter color in the background. The black dots indicate the data collected at five equi-stretch points. Again, these data are generated from the 4FF model ranges in Table~\ref{table:case1_param}. Predictions from the two models match the true data well with slight differences at high-stretch regions. In the right two panels, we plot the stress curves at $\lambda_{z}=1.51$ and $\lambda_{\theta}=1.51$ respectively to compare the GAN-based (red dashed line) and 4FF-based (blue dashed line) mean versus true data (black solid line). The two approaches offer similar prediction accuracy regarding either the mean or standard deviations.

\subsubsection{Inference with partial data}
\label{res_2:partial}

\begin{figure}
    \centering
	\includegraphics[width=1.0\textwidth]{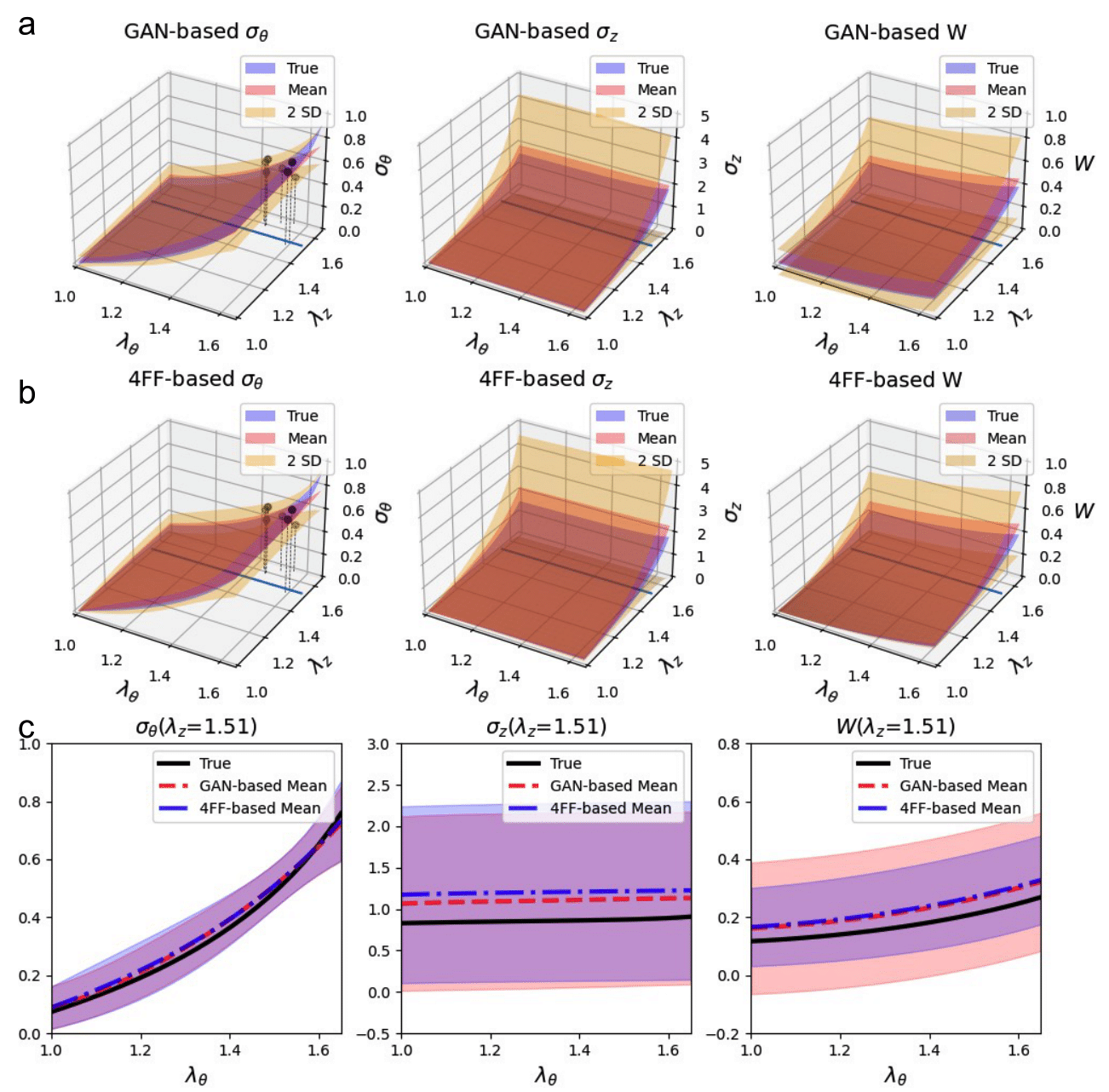} 
    \caption{\textbf{Uninformative inference based on measurements with partial information.} The measurement points only contain information on $\sigma_\theta$ without $\sigma_{z}$. The measurement data are sampled in the range shown in Table~\ref{table:case1_param}. Stress unit: 0.1 MPa.}
    \label{fig:2d_uninform}
\end{figure}

Inferring informative stress/energy fields requires that measurements contain full information in each of the independent directions, e.g., circumferential and axial in this case. Missing certain information will lead to uninformative model inference along the related directions. In this section, we investigate model inference based on partial information, namely, circumferential stress measurements ($\sigma_{\theta}$), without measurements of $\sigma_{z}$. Fig.~\ref{fig:2d_uninform} presents the model inference from the GAN-based and 4FF-based methods with $\sigma_{\theta}$ in the first two rows and compares inference results at $\lambda_{z}=1.51$ in the last row. Estimations of $\sigma_{\theta}$ are accurate and informative for both models, as indicated by accurate mean surfaces (red surface) and relatively small uncertainty bands (orange surface) in the first column. However, both methods lead to large uncertainty bands in predicting $\sigma_{z}$ and energy density.
This result indicates that accurate inference of stresses and energy density fields relies on stress information in both circumferential and axial directions. Otherwise, the inference is not useful as the uncertainty becomes overwhelmingly large regardless of the inference models.

\subsubsection{Inference with Out-Of-Distribution Test}
\label{res_2:ood}
\begin{figure}
    \centering
	\includegraphics[width=0.8\textwidth]{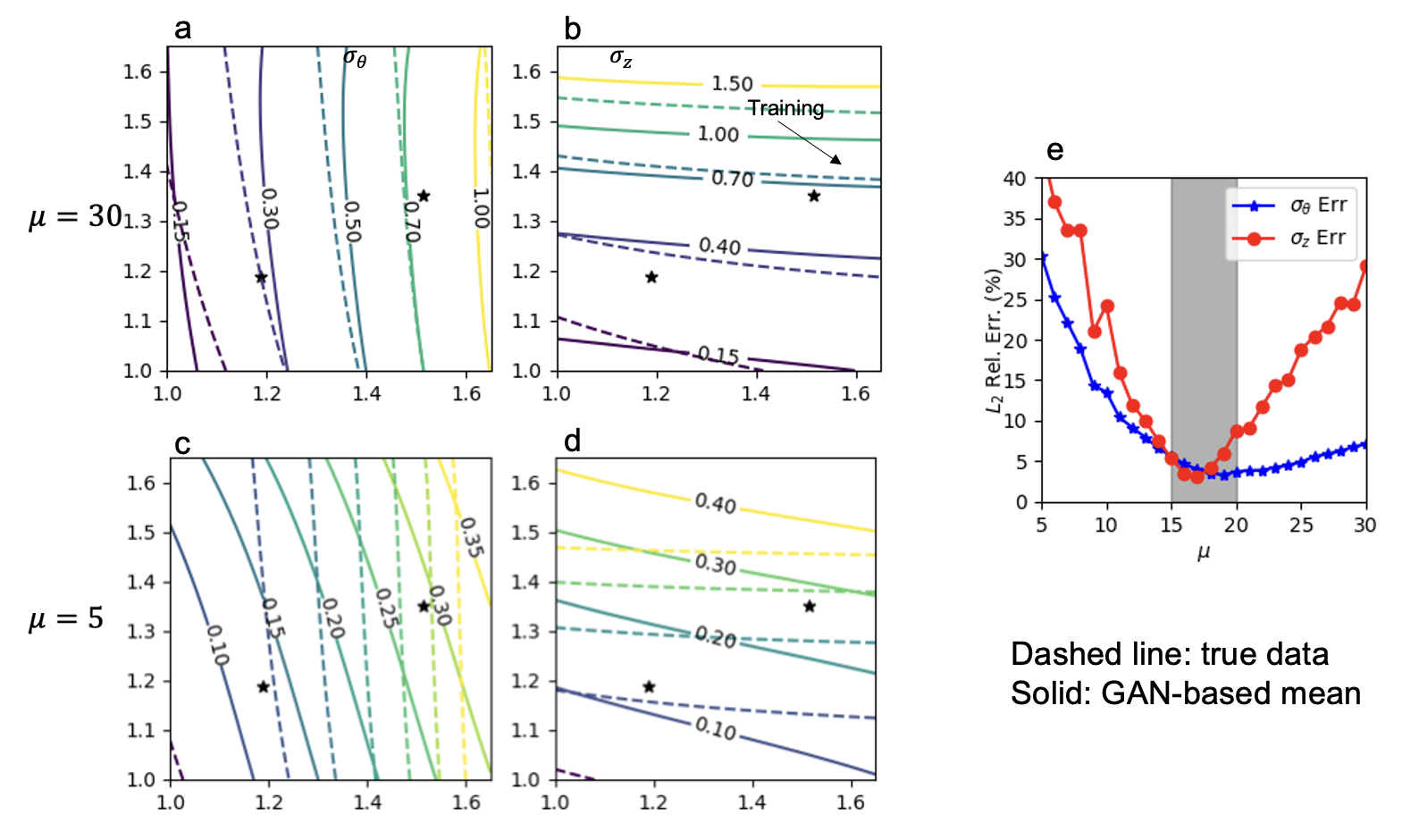} 
    \caption{\textbf{Results of the out-of-distribution (OOD) test.} The GAN-based method is trained on $\mu$ uniformly distributed on $[15.0, 20.0]$ (shaded area) and tested on $\mu\in\{5, 6, ..., 29, 30\}$. $\sigma_{\theta}$ and $\sigma_{z}$ at $\mu=30$ are plotted in a) and b). c) The relative errors are computed between the exact and the predicted mean, estimated from posterior samples, of $\sigma_{\theta}$ and $\sigma_{z}$. Stress unit: 0.1 MPa.}
    \label{fig:ood}
\end{figure}


In this section, we investigate the performance of our model in inferring out-of-distribution (OOD) data, in which there exists a shift in distributions between the training data and the measurement data. In other words, OOD data indicates that the data are drawn from parameters distributed outside of the training data range, often leading to a drop in the model performance. However, the main purpose of testing with OOD data is to see whether our approach is still able to provide relatively accurate inferences as we often encounter scenarios where real-world data and simulated data come from different distributions. First, we choose the prior distribution to be the same as the one defined by Table~\ref{table:case1_param} with the range of shear modulus $\mu$ being set to $[15.0, 20.0]$. The GAN is trained using data randomly sampled from the updated parameter range. The downstream tasks are defined the same as in previous examples, where we collect data measurements $\sigma_{\theta}$ and $\sigma_{z}$ from the 4FF model with $\mu$ sampled from a wider range: $\mu\in\{5, 6, ..., 29, 30\}$. The rest of the parameters are kept inside the training range. Fig.~\ref{fig:ood}(a-d) compares the true contours and model predictions ($\sigma_{\theta}$ and $\sigma_{z}$) at $\mu=30$ and $5$. The measurement data are marked by $*$. It is evident that the predicted contour lines deviate from the true data. In (e), we plot $L_2$ relative errors between the exact and the predicted mean as a function of $\mu$. The errors are relatively small inside the training region (shaded area) $\mu\in[15.0, 20.0]$ but increase as the sampling data deviates from the training region.

\subsection{Case III: 2D Constitutive Law with Ex vivo Data}
\label{res:ex-vivo}

\begin{figure}
    \centering
	\includegraphics[width=1.0\textwidth]{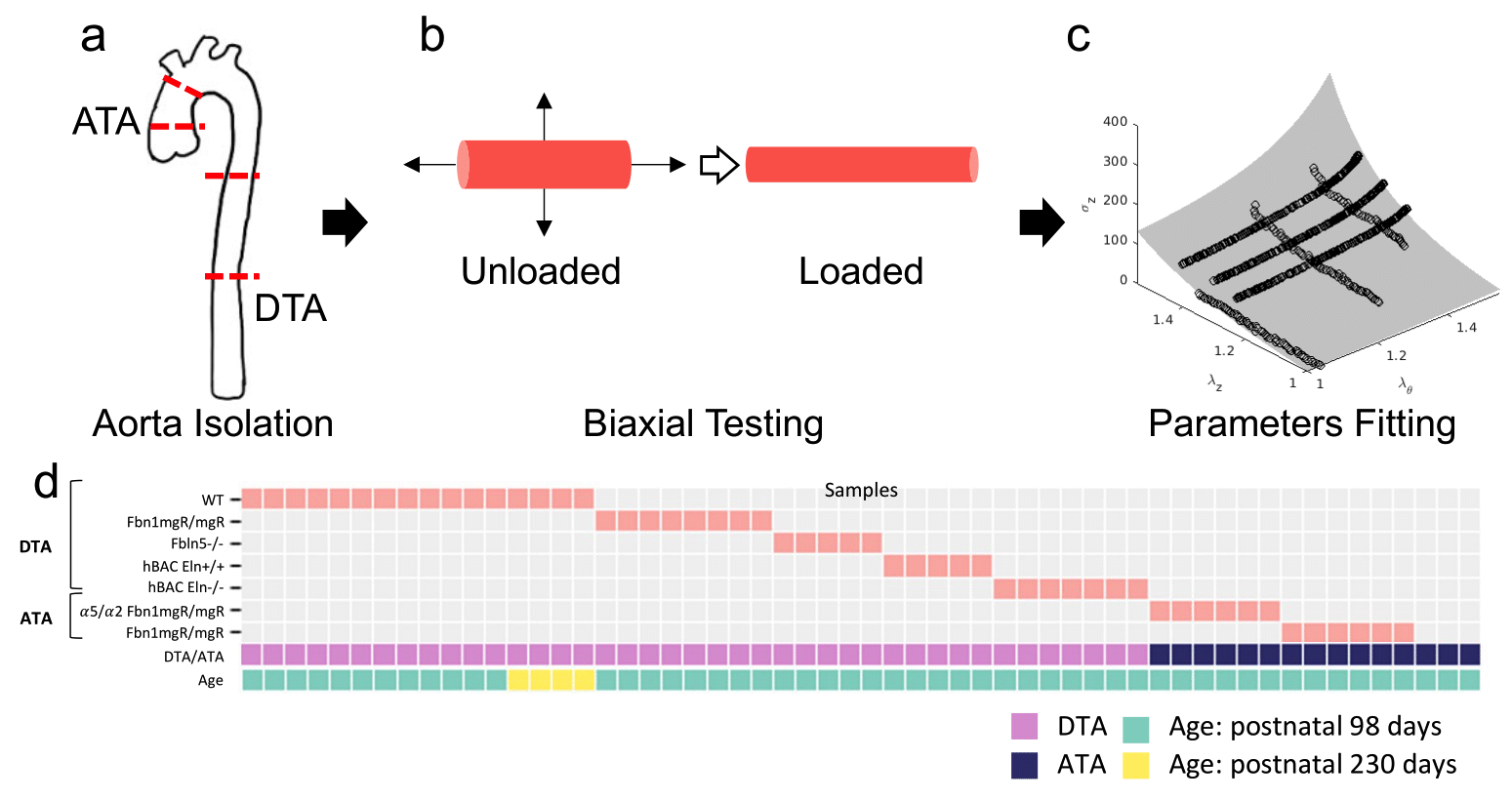} 
    \caption{\textbf{Data acquisition of ex vivo biaxial loading.} (a) Ascending/Descending thoracic aorta (ATA/DTA) from different mice are isolated and tested on (b) biaxially while measuring axial force and pressure. (c) The measured stresses/strains are collected and used in nonlinear parameter fitting.}
    \label{fig:data_acqui}
\end{figure}

In previous sections, we demonstrated our framework's ability in inferring constitutive laws based on synthetic data. Here, we further apply our model to realistic data collected from ex vivo biomechanical experiments for mice with various genotypes. To diversify the dataset, we incorporate data collected from the descending thoracic aorta (DTA) in five mouse models from published studies~\cite{pereira1999pathogenetic,yanagisawa2002fibulin,spronck2021excessive,korneva2019absence,jiao2017deficient,zhang2022g2varphinet} and two more newly-collected groups from the ascending thoracic aorta (ATA). 
\begin{enumerate}
    \item (DTA) Wild-type (WT) C57BL/6J control mice~\cite{jiao2017deficient};
    \item (DTA) $Fbn1^{mgR/mgR}$ mice: normal fibrillin-1 is expressed at 15-25\% of its normal level~\cite{korneva2019absence};
    \item (DTA) $Fbln5^{-/-}$ mice: lack 100\% of the elastin-associated glycoprotein fibulin-5~\cite{ferruzzi2015decreased,spronck2020aortic};
    \item (DTA) hBAC $Eln^{-/-}$ mice: the elastin production was rescued through introduction of human elastin through a bacterial artificial chromosome (hBAC-mNull), resulting in $\sim$30\% of normal elastin expression~\cite{jiao2017deficient};
    \item (DTA) hBAC $Eln^{+/+}$ mice: the elastin is expressed to 115\% of its normal level~\cite{jiao2017deficient};
    \item (ATA) $\alpha$5/$\alpha$2;Fbn1 mgR/mgR: Results of breeding integrin $\alpha$5/$\alpha$2 chimera global knock-in mice on C57BL/6J strain generated using homologous recombination with mgR heterozygous to yield double homozygous mutants;
    \item (ATA) $Fbn1^{mgR/mgR}$ mice: normal fibrillin-1 is expressed at 15-25\% of its normal level~\cite{korneva2019absence};
\end{enumerate}
The stress/stretch data were collected following the procedure of standard biaxial loading tests~\cite{ferruzzi2013biomechanical}, as depicted in Fig.~\ref{fig:data_acqui}. A segment of the DTA or ATA was isolated (Fig.~\ref{fig:data_acqui}(a)),  cannulated, and placed in a biaxial loading device. Next, mechanical loads are collected by the device while pressurizing and stretching the DTA/ATA segment to its in vivo states (Fig.~\ref{fig:data_acqui}(b)). Experimental details of vessel isolation and biaxial loading are elaborated in~\cite{cavinato2021evolving,ferruzzi2013biomechanical}. After data acquisition, we perform a nonlinear regression to identify best-fit values of the 4FF parameters based on all the collected data  (Fig.~\ref{fig:data_acqui}(c)). The 4FF model with fitted parameters renders an overall good approximation of the subject-specific biomechanics with discrepancies at certain stretches. Fig.~\ref{fig:data_acqui}(d) shows the details and diversity of the dataset, e.g., size, genotype, segment location, as well as age, in a heatmap.

\begin{figure}
    \centering
    \subfloat{%
        \centering
        \includegraphics[width=1.0\textwidth]{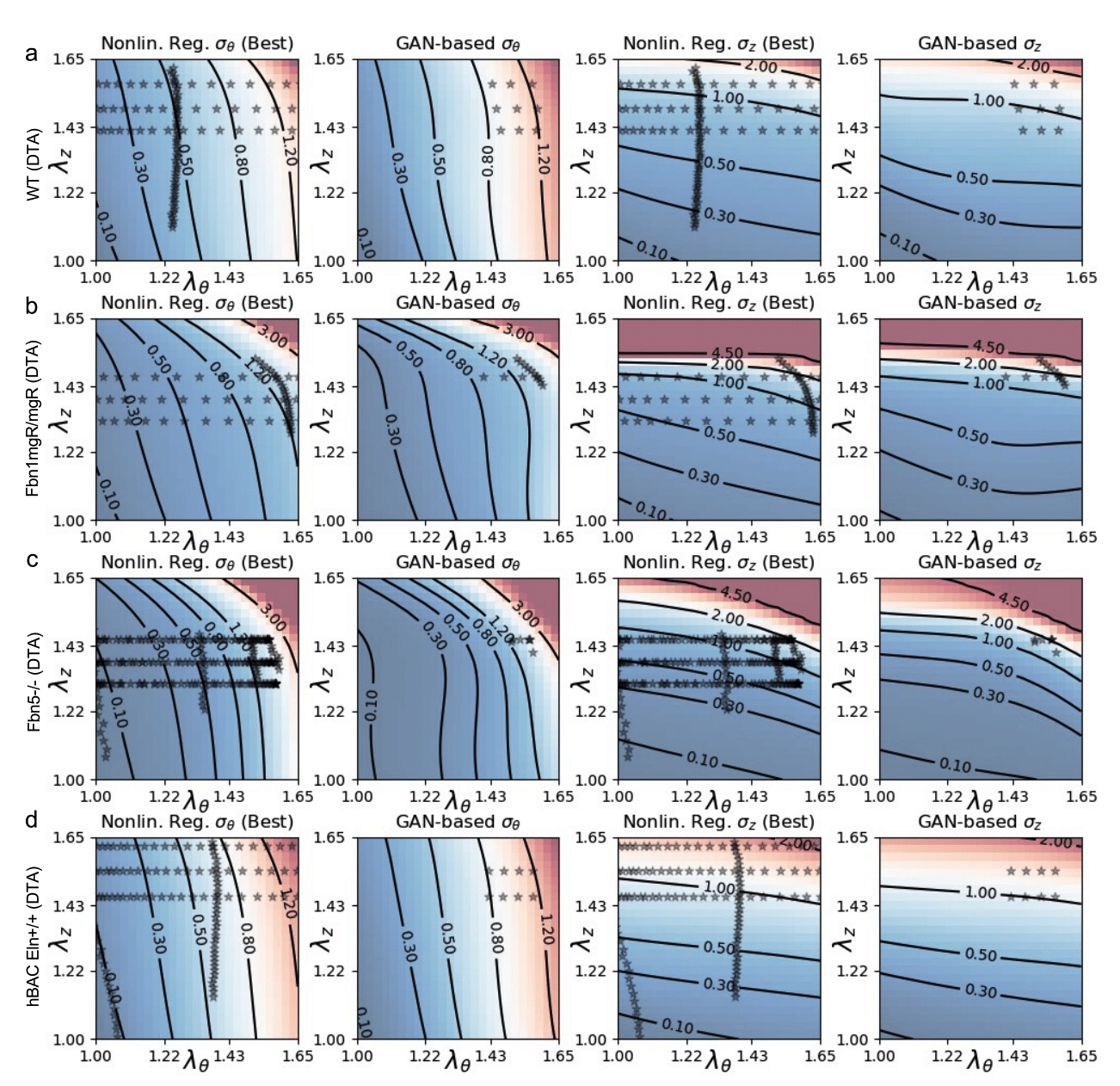}
    }
    \caption{\textbf{(*) Stress inference for DTA isolated from four different genotypes: (a) wild-type, (b) Fbn1mgR/mgR, (c) Fbn5-/-, and (d) hBAC Eln+/+.} The contours of nonlinear regression (Nonlin. Reg.) are calculated with all data collected from ex vivo experiments whereas the GAN-based method estimates the mean of subject-specific constitutive relations only based on sparse data clustered at high stretches. Stress unit: 0.1 MPa.}
    \label{fig:2d_contour_ex_vivo_a}
\end{figure}
    
\begin{figure}
    \ContinuedFloat
    \centering
    \subfloat{%
        \centering
        \includegraphics[width=451pt]{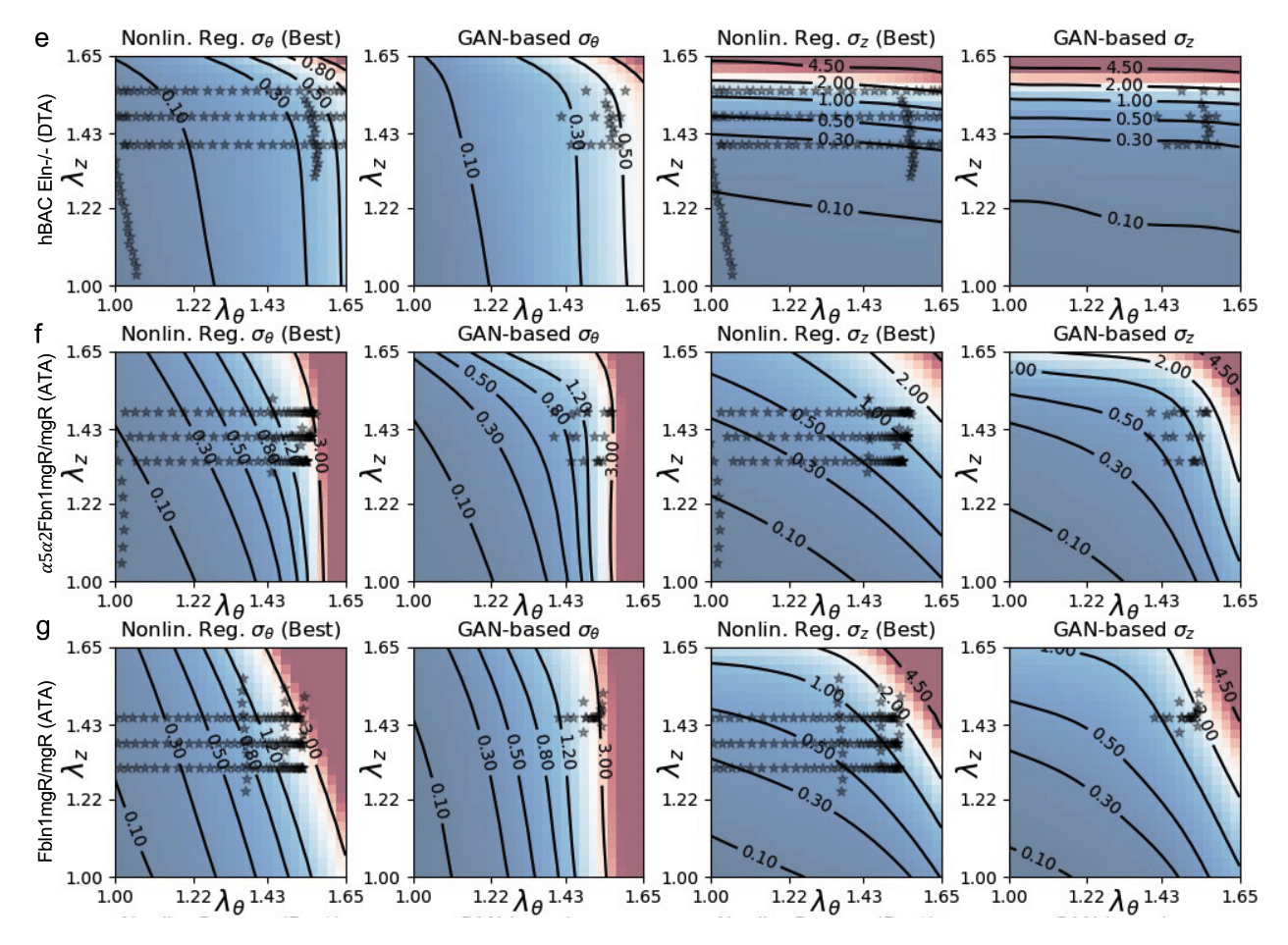}
    }
    \caption{\textbf{(**) Stress inference for three ATA groups and a DTA group: (e) hBAC Eln-/- (DTA), (f) $\alpha$5/$\alpha$2xmgR, and (g) Fbln1 mgR. Stress unit: 0.1 MPa.} Stress unit: 0.1 MPa.}    
    \label{fig:2d_contour_ex_vivo_b}
\end{figure}

Notice that prior knowledge of vascular biomechanics is carried by historical data and informing sufficient data to the generative model is essential for inference accuracy. However, the expensive and time-consuming experimental procedure forbids conducting high-throughput experiments and further restricts learning directly from data. As a proof-of-concept, we adopt data generated from the 4FF model with relatively wide ranges of variations to train the generative model and sample dozens of ex vivo data ($\lambda_{\theta}, \lambda_{z}; \sigma_{\theta}, \sigma_{z}$) in in vivo ranges for posterior estimation. 

Fig.~\ref{fig:2d_contour_ex_vivo_a} presents contours and $\sigma_{\theta}$, $\sigma_z$ fields for different genotypes from nonlinear regression (columns 1 and 3) and the GAN-based method (columns 2 and 4). Each row from (a-h) presents the $\sigma_\theta$ and $\sigma_z$ for a different genotype mouse model with high stresses represented by dark red. The asterisks in each subplot are the data on which the inference is made. In general, results from the GAN-based method indicate that it can efficiently capture the overall trend of stress fields based on sparse measurements (dozens of measurement points) whereas nonlinear regression is performed based on orders of magnitude more measurements. As mentioned, the historical data is generated from the 4FF model in wide variations, which may contain unrealistically high stress at high stretches. To facilitate the network training, we apply a ``cap function'' to stress fields such that stresses greater than 0.5 MPa are truncated at 0.5 MPa. Also, it is interesting to notice some GAN-based results do not satisfy certain properties, e.g., monotonicity, in some regions, such as Fig~\ref{fig:2d_contour_ex_vivo_a}(c) $\sigma_{\theta}=0.1$ or $0.3$.

\begin{figure}
    \centering
	\includegraphics[width=1.0\textwidth]{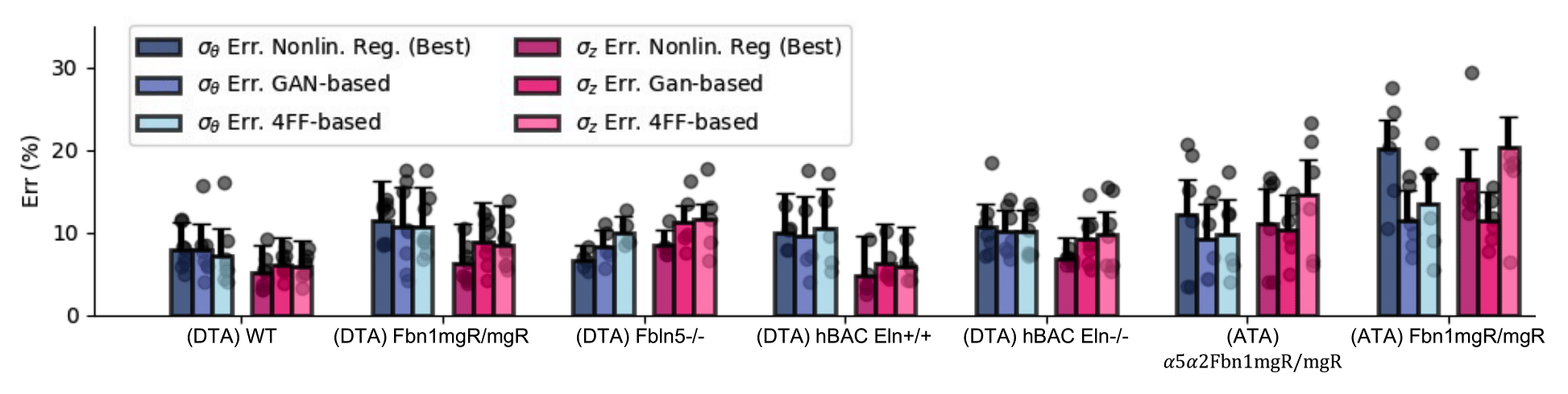} 
    \caption{\textbf{Relative errors of nonlinear regression, GAN-based, and 4FF-based method on data collected from ex vivo experiments.} 
    }
    \label{fig:err_table}
\end{figure}

To quantitatively compare the accuracy between nonlinear regression, GAN-based method, and 4FF-based method, we calculate the mean of relative error based on model predictions at stretches with ex vivo data and plot them in Fig.~\ref{fig:err_table}. Stress errors from different methods are presented in different colors: $\sigma_{\theta}$ errors from nonlinear regression, GAN-based method, and 4FF-based method are plotted in different boxes from dark to light blue; $\sigma_z$ errors are presented in red. The mean errors fluctuate around 10\% across all genotypes for different methods, indicating the similarity in performance achieved by various approaches. In particular, the GAN-based method even outperforms nonlinear regression in $Fbn1^{mgR/mgR}$, possibly due to the latter weighting heavier on large stresses. Overall, we observed a comparable performance between our new method and the standard nonlinear regression.

\section{Discussion}
\label{sec:discussion}

In this paper, we develop a generative modeling framework for inferring families of biomechanical constitutive relations in data-sparse regimes. Inspired by the concept of functional prior~\cite{meng2022learning}, the model is trained with synthetic data from the 4FF model for murine aortas with parameters adopted from published data~\cite{ferruzzi2013biomechanical}. The prior knowledge of the constitutive model is input to a GAN-based network for generating stretch/stress curves and estimating the likelihood function in the subsequent Bayesian inference step. The noise vectors conditioned on sparse measurements are then sampled and transmitted to the generator for inferring the posterior of stresses in the function space. We illustrated the performance of this framework with three examples. The first two examples utilize synthetic data as measurements. The last example demonstrates the model's potential for inferring constitutive relationships based on realistic measurements collected from various murine models. 

We observe that incorporating prior knowledge into neural networks can enhance the accuracy and efficiency of the framework: Prediction errors and standard deviations from our framework are reduced compared to that of GP, which represents model inference without function priors, leading to results prone to be affected by noise and being uninformative outside the sampling range. Unlike linear or nonlinear regressions, the framework is not constrained by certain numerical limitations, e.g., the number of data points must be greater than the number of unknown parameters. This feature shows an advantage over traditional nonlinear regression algorithms for inference with sparse measurements and for complex models with considerable parameters. Moreover, the mean and standard deviations are more informative than traditional algorithms without uncertainty quantification as they represent the expectation and confidence of the prediction.

The advantages of adopting model-agnostic deep learning partially lie in learning systems without explicitly knowing the functional forms. Although standard model-specific approaches could be more efficient as they do not require a network for estimating likelihood for systems with explicit form~\cite{rego2021uncertainty}, it is still worth noticing that our framework demonstrates its advantages when an explicit functional  form is unknown while keeping a similar level of accuracy. Furthermore, model-specific approaches usually require an accurate functional form, whereas a great number of constitutive functions are developed based on certain assumptions and simplifications, e.g., the 4FF model neglects fiber interactions between different alignments. Our framework, however, circumvents this limitation by learning the priors directly from data. Given sufficient training data, our framework has the potential to perform better than model-specific approaches. 


Yet, our generative modeling framework is not without limitations. First, employing such a generative framework is more data-intensive than discriminative models that directly estimate conditional probabilities. Although increasing the training sample size helps to reduce the error (see~\ref{app:B}), it is usually not practical to collect as much data as needed to train such generative models. Another limitation emanated from data acquisition further hurdles training directly from data: the loss of the network is calculated in the whole domain whereas standard biaxial ex vivo measurements cover a fraction of the domain (see $*$ in Fig.~\ref{fig:2d_contour_ex_vivo_a}). 
To mitigate the first limitation, one could attempt to incorporate a hybrid of synthetic and experimental data in the training of generative models, where synthetic data could serve to inform the overall shape of response surfaces while experimental data is used as regularization at certain stretches. Also, encoding physical constraints into the training is a promising direction in generative modeling. Some properties, such as objectivity, consistency, stability, or polyconvexity~\cite{as2022mechanics,linka2023new}, can be utilized as further regularization to help reduce the need for the number of training data and increase accuracy.
As for the second limitation, the unstructured and bias-distributed characteristics of the biaxial ex vivo measurements can also be addressed by incorporating the aforementioned properties into the training processes. Some techniques from the machine learning community to deal with unstructured data \cite{qi2017pointnet, radev2020bayesflow}, e.g. PointNet, may be useful as well. Nevertheless, the aim of this research was not to demonstrate the superiority of either model-specific or model-agnostic approaches as each of them may be suitable in certain circumstances. When the functional  form is available, the model-specific approach is a natural choice. Our contribution focuses on developing a new model-agnostic constitutive modeling that may work when the functional form is unknown and the training data is sufficient.


\section{Acknowledgement}
MY and ZZ would like to thank Professor Xuhui Meng of Huazhong
University of Science and Technology for the insightful discussion. We acknowledge the support by grant U01 HL142518 from the National Institutes of Health.

\appendix


\input{appendix}

\bibliographystyle{ieeetr}
\bibliography{reference}
\end{document}

%% file: appendix.tex
\section{Uniform prior distributions and HMC}\label{app:A}

Hamiltonian Monte Carlo (HMC) is a Markov Chain Monte Carlo (MCMC) method based on Hamiltonian dynamics. It has proven \cite{neal2011mcmc} to be a powerful tool to obtain samples from the posterior distribution, possibly un-normalized, in Bayesian inference, and is widely used to obtain posterior samples of Bayesian neural networks in the presence of noisy measurements \cite{zou2022neuraluq}. As a restriction, HMC requires the density function of the target distribution to be differentiable. However, when a uniform distribution is chosen to be the prior, the posterior distribution $P(\theta|D) \propto P(D|\theta)P(\theta)$ is no longer continuous and differentiable, leading to a difficulty in direct implementation of HMC methods. In this work, we circumvent this limitation by mapping a normal distribution to a uniform distribution
\begin{align}
    X\sim \mathcal{N}(0, 1^2) \longrightarrow a + (b-a) F_X(X) \sim U(a, b),
\end{align}
where $X$ is a random variable subject to Gaussian distribution with $0$ mean and $1$ standard deviation, $a<b$ are real values, and $F_X$ is the cumulative distribution function of a standard normal random variable, i.e. $F_X(x) := \int_{-\infty}^x \frac{1}{\sqrt{2\pi}}\exp(-\frac{s^2}{2})ds$. Hence, for each model parameter with a uniform prior distribution, $\theta\sim U(a, b)$, there exists a latent variable $X_\theta\sim \mathcal{N}(0, 1^2)$ such that $\theta$ and $a + (b-a)F_X(X_\theta)$ are equal in distribution, and therefore HMC is instead done on the latent variables, whose joint prior distribution has differentiable density function.

\section{The effect of the size of historical data}
\label{app:B}

Recall that a GAN was trained on historical data of $\sigma_\theta$ and $\sigma_z$, $\tilde{T} = \{T_i\}_{i=1}^N$, generated from the 4FF model, where each $T_i$ is a finite representation of a model from the target family and $N$ is the number of data. The accuracy of GANs in learning 4FF model is therefore crucial for accurate and trustworthy prediction in the downstream tasks. 
We conduct a simple experiment in this section, demonstrating the number of historical data used for training GANs and their effect on the downstream inferences.

\begin{figure}
    \centering
    \includegraphics[width = 1.0\textwidth]{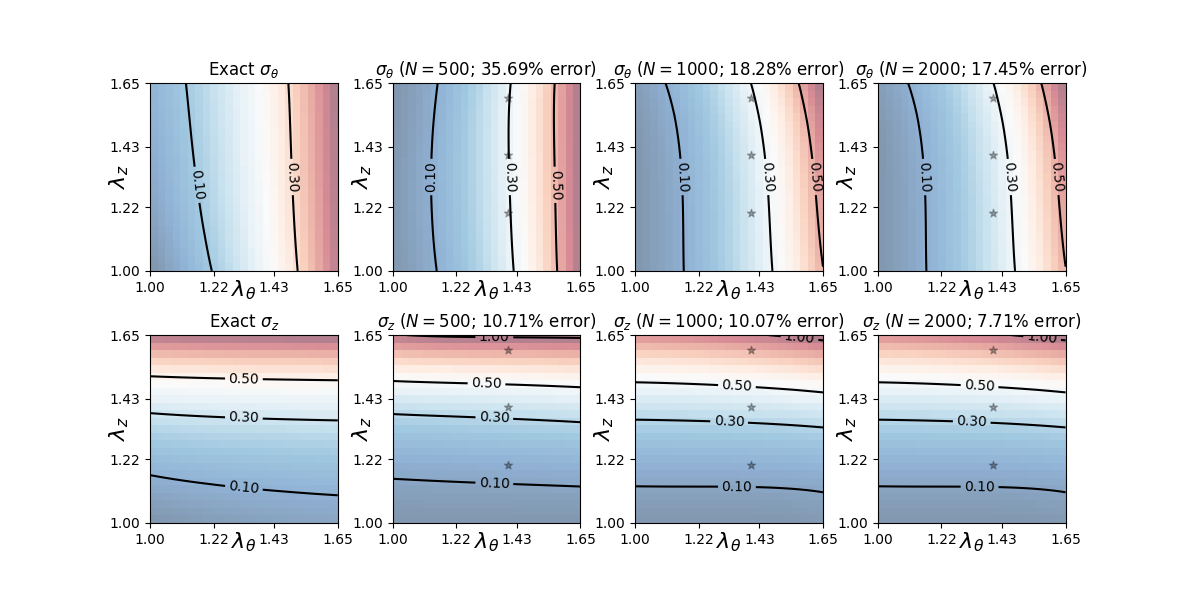}
    \caption{\textbf{Results of downstream inferences from sparse measurements (simulated data) with GANs trained on different numbers of data.} The number of data refers to the case number used to train GAN-based priors. The accuracy of the inference increases with the number of training data. The $L_2$ relative errors of the inferences are shown at the top of each figure. Violation of the constitutive laws also appears in inferring the circumferential stress $\sigma_\theta$ using the GAN-based prior trained on $N=500$ data. Stress unit: 0.1 MPa.}
    \label{fig:appendix_2d_contour}
\end{figure}

Here we use three datasets ($N=500$, $N=1,000$, and $N=2,000$) for GAN training. In the downstream tasks, we test model accuracy based on the same measurements, $D$. The noise is additive Gaussian with zero mean and $0.1$ standard deviation. The distribution of the 4FF model is taken from Table~\ref{table:case1_param}. Results are shown in Fig.~\ref{fig:appendix_2d_contour}, from which we can see that the accuracy and robustness increase with the number of training data. $L_2$ relative errors of the predicted means are 35.69\% from 500 data, 18.28\% from 1000 data, 17.45\% from 2000 data, for circumferential stress, and 10.71\%, 10.07\%, 7.71\% for axial stress.
Besides, we also notice that the circumferential stress violates monotonicity in the 500 data case (see the $\lambda_\theta \in [1.00, 1.22]$ region and $\sigma_\theta=0.1$ contour line in the first row and second column of Fig.~\ref{fig:appendix_2d_contour}), indicating that prior knowledge is not well-captured by the network due to insufficient training data.

\section{The effect of sampling regions}\label{app:C}

\begin{figure}
    \centering
    \includegraphics[width = 1.0\textwidth]{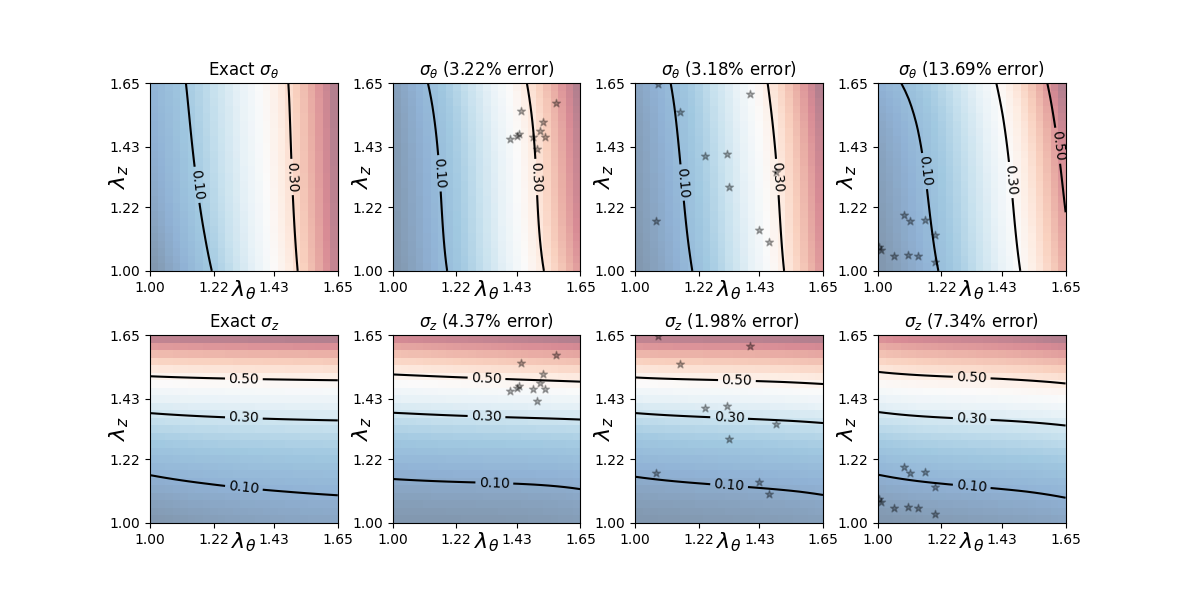}
    \caption{\textbf{Results of downstream inferences from measurements sampled in various regions.} The number of measurements is set to be $20$ across all three cases. The measurements are assumed to be clean and their locations are randomly sampled in $[1.4, 1.6]^2$, $[1.0, 1.65]^2$ (the whole domain of interest), and $[1.0, 1.2]^2$, respectively. The $L_2$ relative errors of the inferences are shown at the top of each figure. Among all three, the spreading sampling technique delivers the most accurate inferences. Stress unit: 0.1 MPa}
    \label{fig:appendix_2d_contour_2}
\end{figure}

In this section, we investigate the effects of measurement locations on the accuracy of downstream inferences. 
With the same trained network, we conduct inference using the same number of measurements sampled at different regions. 
As shown in Fig.~\ref{fig:appendix_2d_contour_2}, the sampling locations are clustered in three different ways: in high-stretch $\lambda_{z}\times\lambda_\theta\in[1.4, 1.6]^2$, in low-stretch $\lambda_{z}\times\lambda_\theta\in[1.0, 1.2]^2$, and spreading over the domain of interest, $\lambda_{z}\times\lambda_\theta\in[1, 1.65]^2$. Among all three cases, the inferences using data spreading the whole domain are the most accurate whereas the case where data is clustering in $[1.0, 1.2]^2$ produces the worst results. High-stretch clustering performs a bit worse than spreading sampling.

